\documentclass[11pt,a4paper]{article}

\pdfoutput=1

\usepackage[hyperref]{acl2017}
\usepackage{times}
\usepackage{latexsym}
\usepackage{amsmath}
\usepackage{amsthm}
\usepackage{framed}
\usepackage{multirow}
\usepackage{caption}
\usepackage{subcaption}
\usepackage{latexsym}
\usepackage{url}
\usepackage{graphicx}
\usepackage{xcolor,colortbl}

\aclfinalcopy 

\definecolor{Gray}{gray}{0.85}
\definecolor{lGray}{gray}{0.95}

\title{Lexical Features in Coreference Resolution: To be Used With Caution}

\author{Nafise Sadat Moosavi \and Michael Strube\\
  Heidelberg Institute for Theoretical Studies gGmbH \\
  Schloss-Wolfsbrunnenweg 35 \\
  69118 Heidelberg, Germany \\
  {\small \url{{nafise.moosavi|michael.strube}@h-its.org}}}

\date{}

\begin{document}
\maketitle
\begin{abstract}
	Lexical features are a major source of information in state-of-the-art coreference resolvers.
	Lexical features implicitly model some of the linguistic phenomena at a fine granularity level.
	They are especially useful for representing the context of mentions.
	In this paper we investigate a drawback of using many lexical features in  
	state-of-the-art coreference resolvers.
	We show that if coreference resolvers mainly rely on lexical features, they can hardly generalize to unseen domains.
Furthermore, we show that the current coreference resolution evaluation is clearly flawed by only evaluating on a specific split of a specific dataset
in which there is a notable overlap between the training, development and test sets.

\end{abstract}
\section{Introduction}
\label{sect:intro}
Similar to many other tasks, lexical features are a major source of information in current coreference resolvers.
Coreference resolution is a set partitioning problem in which each resulting partition refers to an entity.
As shown by \newcite{durrett13b}, lexical features implicitly model some linguistic phenomena, which were previously modeled by heuristic features, but at a finer level of granularity.
However, we question whether the knowledge that 
is mainly captured by lexical features can be generalized to other domains.   

The introduction of the CoNLL dataset enabled a significant boost in the performance of coreference resolvers,
i.e.\ about 10 percent difference between the CoNLL score of the currently best coreference resolver, deep-coref by \newcite{clarkkevin16a}, 
and the winner of the CoNLL 2011 shared task, the Stanford rule-based system by \newcite{leeheeyoung13}. 
However, this substantial improvement does not seem to be visible in downstream tasks.
Worse, the difference between state-of-the-art coreference resolvers and the rule-based system drops significantly 
when they are applied on a new dataset, 
even with consistent definitions of mentions and coreference relations \cite{ghaddar16a}.
%

In this paper, we show that if we mainly rely on lexical features,
as it is the case in state-of-the-art coreference resolvers,
overfitting become more sever. 
Overfitting to the training dataset is a problem that cannot be completely avoided.
However, there is a notable overlap between the CoNLL training, development and test sets
that encourages overfitting.
Therefore, the current coreference evaluation scheme is flawed by only evaluating on this overlapped validation set.
To ensure meaningful improvements in coreference resolution,
we believe an out-of-domain evaluation is a must in the coreference literature.
%
%
\section{Lexical Features}
\label{sect:benefit}
\newcolumntype{a}{>{\columncolor{lGray}}r}
\newcolumntype{b}{>{\columncolor{Gray}}r}
\begin{table*}[t]
    \begin{center}\footnotesize
    \resizebox{2\columnwidth}{!}{%
    \begin{tabular}{l|rra|rra|rra|b||rra}
     \multicolumn{1}{c}{} & \multicolumn{3}{c}{MUC} &
     \multicolumn{3}{c}{$B^3$} & \multicolumn{3}{c}{CEAF$_e$} & \multicolumn{1}{c}{CoNLL} & \multicolumn{3}{c}{LEA} \\ \hline
     \multicolumn{1}{c|}{}&R & P & F$_1$ & R & P & F$_1$ & R & P & F$_1$ & Avg. F$_1$ & R & P & F$_1$\\ \hline
         \multicolumn{1}{c|}{} & \multicolumn{13}{c}{CoNLL test set} \\ \hline
     \hline
     rule-based & $64.29$ & $65.19$ & $64.74$ & $49.18$ & $56.79$ & $52.71$ & $52.45$ & $46.58$ & $49.34$ & $55.60$ & $43.72$ & $51.53$ & $47.30$ \\
	 berkeley & $67.56$ & $74.09$ & $70.67$ & $53.93$ & $63.50$ & $58.33$ & $53.29$ & $56.22$ & $54.72$ & $61.24$ & $49.66$ & $59.17$ & $54.00$ \\
     cort  & $67.83$ & $78.35$ & $72.71$ & $54.34$ & $68.42$ & $60.57$ & $53.10$ & $61.10$ & $56.82$ & $63.37$ & $50.40$ & $64.46$ &$56.57$\\ 
     deep-coref [conll] & $70.55$ & $79.13$ & $74.59$ & $58.17$ & $69.01$ & $63.13$ & $54.20$ & $63.44$ & $58.45$ & $65.39$ & $54.55$ & $65.35$ & $59.46$ \\
     deep-coref [lea] & $70.43$ & $79.57$ & $74.72$ & $58.08$ & $69.26$ & $63.18$ & $54.43$ & $64.17$ & $58.90$ & $65.60$ & $54.55$ & $65.68$ & $59.60$ \\
	  \hline
	  \multicolumn{1}{c}{} & \multicolumn{13}{c}{WikiCoref} \\ \hline
      rule-based  & $60.42$ & $61.56$ & $60.99$ & $43.34$ & $53.53$ & $47.90$ & $50.89$ & $42.70$ & $46.44$ & $51.77$ & $38.79$ & $48.92$ & $43.27$ \\
      berkeley & $68.52$ & $55.96$ & $61.61$ & $59.08$ & $39.72$ & $47.51$ & $48.06$ & $40.44$ & $43.92$ & $51.01$ & - & - & - \\
      cort & $70.39$ & $53.63$ & $60.88$ & $60.81$ & $37.58$ & $46.45$ & $47.88$ & $38.18$ & $42.48$ & $49.94$ & - & - & -\\
      deep-coref [conll] & $58.59$ & $66.63$ & $62.35$ & $44.40$ & $54.87$ & $49.08$ & $42.47$ & $51.47$ & $46.54$ & $52.65$ & $40.36$ & $50.73$ & $44.95$\\
      deep-coref [lea]& $57.48$ & $70.55$ & $63.35$ & $42.12$ & $60.13$ & $49.54$ & $41.40$ & $53.08$ & $46.52$ & $53.14$ & $38.22$ & $55.98$ & $45.43$\\
      deep-coref$^-$ & $55.07$ & $71.81$ & $62.33$ & $38.05$ & $61.82$ & $47.11$ &  $38.46$ & $50.31$ & $43.60$ & $51.01$ & $34.11$ & $57.15$ & $42.72$ \\
    \end{tabular}
    }
    \end{center}
    \caption{Comparison of the results on the CoNLL test set and WikiCoref.}
    \label{tab:conll_data}
\end{table*}
The large difference in performance between coreference resolvers that use lexical features and ones which do not,
implies the importance of lexical features.
\newcite{durrett13b} show that lexical features implicitly capture some phenomena, e.g.\ definiteness and syntactic roles, which were previously modeled by heuristic features.
\newcite{durrett13b} use exact surface forms as lexical features.
However, when word embeddings are used instead of surface forms,
the use of lexical features is even more beneficial.
Word embeddings are an efficient way of capturing semantic relatedness.
Especially, they provide an efficient way for describing the context of mentions.

\newcite{durrett13b} show that the addition of some heuristic features like gender, number, person and animacy agreements and syntactic roles on top of their lexical features 
does not result in a significant improvement.

deep-coref, the state-of-the-art coreference resolver, follows the same approach.
\newcite{clarkkevin16a} capture the required information for resolving coreference relations by using 
a large number of lexical features
and a small set of non-lexical features including string match, distance, mention type, speaker and genre features.
The main difference is that \newcite{clarkkevin16a} use word embeddings instead of the exact surface forms that are used by \newcite{durrett13b}.

Based on the error analysis by cort \cite{martschat14},
in comparison to systems that do not use word embeddings, 
deep-coref has fewer recall and precision errors especially for pronouns.  
For example, deep-coref correctly recognizes around $83$ percent of non-anaphoric ``it'' in the CoNLL development set. This could be a direct result of a better context representation by word embeddings. 
\section{Out-of-Domain Evaluation}
\label{sect:out-of-domain}
Aside from the evident success of lexical features, 
it is debatable how well the knowledge that is mainly captured by the lexical information of the training data
can be generalized to other domains.
As reported by \newcite{ghaddar16b},
state-of-the-art coreference resolvers trained on the CoNLL dataset perform poorly, i.e.\ worse than the rule-based system \cite{leeheeyoung13}, 
on the new dataset, WikiCoref \cite{ghaddar16b},
even though WikiCoref is annotated with the same annotation guidelines as the CoNLL dataset.
The results of some of recent coreference resolvers on this dataset are listed in Table~\ref{tab:conll_data}.

The results are reported using \emph{MUC} \cite{vilain95}, \emph{B}$^3$ \cite{bagga98b}, \emph{CEAF}$_e$ \cite{luoxiaoqiang05a}, 
the average $F_1$ score of these three metrics, i.e.\ CoNLL score, and \emph{LEA} \cite{moosavi16b}.

\emph{berkeley} is the mention-ranking model of \newcite{durrett13b} with the FINAL feature set including
the head, first, last, preceding and following words of a mention, 
the ancestry, length, gender and number of a mention, distance of two mentions, 
whether the anaphor and antecedent are nested, same speaker and a small set of string match features.

\emph{cort} is the mention-ranking model of \newcite{martschat15c}.
\emph{cort} uses the following set of features: 
the head, first, last, preceding and following words of a mention, 
the ancestry, length, gender, number, type, semantic class, dependency relation and dependency governor of a mention,
the named entity type of the head word, distance of two mentions, 
same speaker, whether the anaphor and antecedent are nested, and a set of string match features.
\emph{berkeley} and \emph{cort} scores in Table~\ref{tab:conll_data} are taken from \newcite{ghaddar16a}.

\emph{deep-coref} is the mention-ranking model of \newcite{clarkkevin16a}.
\emph{deep-coref} incorporates a large set of embeddings, i.e.\ 
embeddings of the head, first, last, two previous/following words, and the dependency governor of a mention
in addition to the averaged embeddings of the five previous/following words, all words of the mention, sentence words, and document words. 
\emph{deep-coref} also incorporates type, length, and position of a mention, whether the mention is nested in any other mention, distance of two mentions, speaker features and a small set of string match features. 
 
For \emph{deep-coref [conll]} the averaged CoNLL score is used to select the best trained model on the development set.
\emph{deep-coref [lea]} uses the \emph{LEA} metric \cite{moosavi16b} for choosing the best model.
It is worth noting that the results of \emph{deep-coref}'s ranking model may be slightly different at various experiments.
However, the performance of \emph{deep-coref [lea]} is always higher than that of \emph{deep-coref [conll]}. 

We add WikiCoref's words to \emph{deep-coref}'s dictionary for both \emph{deep-coref [conll]} and \emph{deep-coref [lea]}.
\emph{deep-coref}$^-$ reports the performance of \emph{deep-coref [lea]} in which WikiCoref's words are not incorporated into the dictionary.
Therefore, for \emph{deep-coref}$^-$, WikiCoref's words that do not exist in CoNLL will be initialized randomly instead of using pre-trained \emph{word2vec} word embeddings.
The performance gain of \emph{deep-coref [lea]} in comparison to \emph{deep-coref}$^-$ 
indicates the benefit of using pre-trained word embeddings and word embeddings in general. 
Henceforth, we refer to \emph{deep-coref [lea]} as \emph{deep-coref}.

\section{Why do Improvements Fade Away?}
\label{sect:barrier}
\begin{table}[t]
    \begin{center}\footnotesize
    \begin{tabular}{rrrrrrr}
		\multicolumn{7}{c}{genre} \\ \hline
     bc & bn & mz & nw & pt & tc & wb \\ \hline
     \multicolumn{7}{c}{train+dev} \\ \hline
     $43\%$ & $50\%$ & $51\%$ & $45\%$ & $77\%$ & $38\%$ & $39\%$\\ \hline
      \multicolumn{7}{c}{train} \\ \hline
     $41\%$ & $49\%$ & $39\%$ & $44\%$ & $76\%$ & $37\%$ & $38\%$\\
    \end{tabular}
    \end{center}
    \caption{Ratio of non-pronominal coreferent mentions in the test set that are seen as coreferent in the training data.}
    \label{tab:coref_ratio}
\end{table}
\begin{table*}[t]
    \begin{center}\footnotesize
    \begin{tabular}{l|r|rrr||r|rrr}
     \multicolumn{1}{c}{} & CoNLL & \multicolumn{3}{c||}{LEA} & CoNLL & \multicolumn{3}{c}{LEA}  \\ \hline
     \multicolumn{1}{c|}{}& Avg. F$_1$ & R & P & F$_1$& Avg. F$_1$ & R & P & F$_1$\\ \hline
     \multicolumn{1}{c}{} & \multicolumn{8}{c}{pt} \\ \hline
     \multicolumn{1}{c|}{} & \multicolumn{4}{c||}{in-domain} & \multicolumn{4}{c}{out-of-domain} \\ \hline
     rule-based & - & - & - & - & $65.01$ & $50.58$ & $65.02$ & $56.90$\\
     berkeley-surface & $69.15$ & $58.57$ & $65.24$ & $61.73$ & $63.01$ & $46.56$ & $62.13$ & $53.23$ \\
	 berkeley-final & $70.71$ & $60.48$ & $67.29$ & $63.70$ & $64.24$ & $47.10$ & $65.77$ & $54.89$\\
     cort & $72.56$ & $61.82$ & $70.70$ & $65.96$ & $64.60$ & $46.85$ & $67.69$ & $55.37$ \\
     cort$-$lexical & $69.48$ & $54.26$ & $70.33$ & $61.26$ & $64.32$ & $45.63$ & $68.51$ & $54.77$\\
     deep-coref & $75.61$ & $68.48$ & $73.70$ & $71.00$ & $66.06$ & $52.44$ & $63.84$ & $57.58$  \\ \hline
     \multicolumn{1}{c}{} & \multicolumn{8}{c}{wb} \\ \hline
     \multicolumn{1}{c|}{} & \multicolumn{4}{c||}{in-domain} & \multicolumn{4}{c}{out-of-domain} \\ \hline
     rule-based & - & - & - & - & $53.80$ & $45.19$ & $44.98$ & $45.08$\\
     berkeley-surface & $56.37$ & $45.72$ & $47.20$ & $46.45$ & $55.14$ & $45.94$ & $44.59$ & $45.26$\\
	 berkeley-final & $56.08$ & $44.20$ & $50.45$ & $47.12$ & $57.31$ & $50.33$ & $46.17$ & $48.16$\\
     cort & $59.29$ & $50.37$ & $51.56$ & $50.96$ & $58.87$ & $51.47$ & $50.96$ & $51.21$ \\
     cort$-$lexical & $56.83$ & $51.00$ & $47.34$ & $49.10$ & $57.10$ & $51.50$ & $47.83$ & $49.60$ \\
     deep-coref & $61.46$ & $48.04$ & $60.99$ & $53.75$ & $57.17$ & $50.29$ & $47.27$ & $48.74$  \\ \hline
    \end{tabular}
    \end{center}
    \caption{In-domain and out-of-domain evaluations for a high and a low overlapped genres.}
    \label{tab:cross_genre}
\end{table*}
%
In this section, we investigate how much lexical features contribute to the fact that 
current improvements in coreference resolution do not properly apply to a new domain.

Table~\ref{tab:coref_ratio} shows the ratio of non-pronominal coreferent mentions in the CoNLL test set that also appear as coreferent mentions in the training data.   
These high ratios indicate a high degree of overlap between the mentions 
of the CoNLL datasets.

The highest overlap between the training and test sets exists in genre \emph{pt} (Bible).
The \emph{tc} (telephone conversation) genre has the lowest overlap for non-pronominal mentions.
However, this genre includes a large number of pronouns.
We choose \emph{wb} (weblog) and \emph{pt} for our analysis as two genres with low and high degree of overlap.

Table~\ref{tab:cross_genre} shows the results of the examined coreference resolvers when the test set only includes one genre, i.e.\ \emph{pt} or \emph{wb}, in two different settings: (1)
the training set includes all genres (in-domain evaluation), and (2) the corresponding genre of the test set is excluded from the training and development sets (out-of-domain evaluation).

\emph{berkeley-final} is the coreference resolver of \newcite{durrett13b} with the FINAL feature set explained in Section~\ref{sect:out-of-domain}.
\emph{berkeley-surface} is the same coreference resolver with only surface features, i.e.\ ancestry, gender, number, same speaker and nested features are excluded from the FINAL feature set.

\emph{cort$-$lexical} is a version of \emph{cort} in which no lexical feature is used, i.e.\ 
the head, first, last, governor, preceding and following words of a mention are excluded.

For in-domain evaluations we train \emph{deep-coref}'s ranking model for 100 iterations, 
i.e. the setting used by \newcite{clarkkevin16b}. 
However, based on the performance on the development set, we only train the model for 50 iterations in out-of-domain evaluations.  

The results of the \emph{pt} genre show that when there is a high overlap between the training and test datasets, the performance of all learning-based classifiers significantly improves.
\emph{deep-coref} has the largest gain from including \emph{pt} in the training data that is more than 13\% based on the \emph{LEA} score.
\emph{cort} uses both lexical and a relatively large number of non-lexical features
while \emph{berkeley-surface} is a pure lexicalized system.
However, the difference between the \emph{berkeley-surface}'s performances when \emph{pt} is included or excluded from the training data is lower than that of \emph{cort}.
\emph{berkeley} uses feature-value pruning so lexical features that occur fewer than 20 times are pruned from the training data.
Maybe, this is the reason that \emph{berkeley}'s performance difference is less than other lexicalized systems in highly overlapping datasets. 

For a less overlapping genre, i.e.\ \emph{wb}, the performance gain of including the genre in the training data  is significantly lower for all lexicalized systems. 
Interestingly, the performance of \emph{berkeley-final}, \emph{cort} and \emph{cort$-$lexical} increases for the \emph{wb} genre when this genre is excluded from the training set.
 \emph{deep-coref}, which uses a complex deep neural network and mainly lexical features, has the highest gain from the redundancy in the training and test datasets.
As we use more complex neural networks, there is more capacity for brute-force memorization of the training dataset.

It is also worth noting that the performance gains and drops in out-of-domain evaluations are not entirely because of lexical features, as the performance of \emph{cort$-$lexical} also drops significantly in \emph{pt} out-of-domain evaluation.
The classifier may also memorize other properties of the seen mentions in the training data.
However, in comparison to features like gender and number agreement or syntactic roles, lexical features have the highest potential for overfitting.  

%
%
\begin{table}[t]
    \begin{center}\footnotesize
    \begin{tabular}{ll|rrr}
	 \multicolumn{2}{c}{} &	\multicolumn{3}{c}{Anaphor} \\
     \multicolumn{1}{c}{Antecedent} & & \multicolumn{1}{c}{Proper} & \multicolumn{1}{c}{Nominal} & \multicolumn{1}{c}{Pronominal} \\ \hline
	 \multirow{2}{*}{Proper} 
	 &seen & $80\%$& $85\%$ & $77\%$\\
	 &all &$3221$ & $261$& $1200$\\
	  \hline
	 \multirow{2}{*}{Nominal} 
	 &seen & $75\%$ &$93\%$ & $95\%$\\
	 &all & $69$ & $1673$ & $1315$\\
	  \hline
	 \multirow{2}{*}{Pronominal}
	  & seen &$58\%$ &$99\%$ &$100\%$ \\
	  & all & $85$ &$74$&$4737$ \\
	 \hline
    \end{tabular}
    \end{center}
    \caption{Ratio of links created by {deep-coref} for which the head-pair is seen in the training data.}
    \label{tab:deep_coref_head_seen}
\end{table}
\begin{table}[t]
    \begin{center}\footnotesize
    \begin{tabular}{ll|rrr}
	 \multicolumn{2}{c}{} &	\multicolumn{3}{c}{Anaphor} \\
     \multicolumn{1}{c}{}& & \multicolumn{1}{c}{Proper} & \multicolumn{1}{c}{Nominal} & \multicolumn{1}{c}{Pronominal} \\ \hline
	 \multicolumn{1}{c}{Antecedent} & \multicolumn{3}{c}{Correct decisions} \\ \hline
	 \multirow{2}{*}{Proper} 
	 & seen & $82\%$ &$85\%$ & $78\%$\\
	 & all & $2603$ & $150$ & $921$\\
	 \hline
	 \multirow{2}{*}{Nominal} 
	  & seen & $76\%$ & $94\%$ & $96\%$\\
	  & all & $42$ & $1058$ & $890$\\ \hline
	 \multirow{2}{*}{Pronominal}
	 & seen & $63\%$ & $98\%$ & $100\%$\\
	 & all & $49$ & $44$ & $3998$ \\
	  
	 \hline
	 \multicolumn{1}{c}{} & \multicolumn{3}{c}{Incorrect decisions} \\ \hline
	 \multirow{2}{*}{Proper} 
     & seen & $73\%$ & $85\%$ & $76\%$\\
	 & a11 & $618$ & $111$ & $279$\\ \hline
	 \multirow{2}{*}{Nominal}
	 & sen & $74\%$ & $92\%$ & $94\%$\\
	 &all & $27$ & $615$ & $425$\\
	  \hline
	 \multirow{2}{*}{Pronominal} 
	 & seen & $50\%$ & $100\%$ & $100\%$\\
	 & all & $36$ & $30$ &$739$\\
	 \hline
    \end{tabular}
    \end{center}
    \caption{Ratio of links created by {deep-coref} for which the head-pair is seen in the training data.}
    \label{tab:deep_coref_head_seen_correct}
\end{table}
\begin{table}[t]
    \begin{center}\footnotesize
    \begin{tabular}{ll|rrr}
	 \multicolumn{2}{c}{} &	\multicolumn{3}{c}{Anaphor} \\
     \multicolumn{1}{c}{Antecedent} & & \multicolumn{1}{c}{Proper} & \multicolumn{1}{c}{Nominal} & \multicolumn{1}{c}{Pronominal} \\ \hline
	 \multirow{2}{*}{Proper} 
	 &seen &  63\% & 51\% & 75\%\\ 
	 &all &  818 & 418 & 278 \\ 
	  \hline
	 \multirow{2}{*}{Nominal} 
	 &seen &  44\% & 73\% & 90\%\\
	 &all &  168 & 892 & 538\\
	  \hline
	 \multirow{2}{*}{Pronominal}
	  & seen & 82\% & 90\% & 100\%\\
	  & all & 49 & 59 & 444\\
	 \hline
    \end{tabular}
    \end{center}
    \caption{Ratio of deep-coref's recall errors for which the head-pair exists in the training data.}
    \label{tab:deep_coref_recall}
\end{table}

We further analyze the output of \emph{deep-coref} on the development set.
The \emph{all} rows in Table~\ref{tab:deep_coref_head_seen} show the number of pairwise links that are created by \emph{deep-coref} on the development set 
for different mention types.
The \emph{seen} rows show the ratio of each category of links for which the (antecedent head, anaphor head) pair is seen in the training set.
All ratios are surprisingly high. 
The most worrisome cases are those in which both mentions are either a proper name or a common noun.

Table~\ref{tab:deep_coref_head_seen_correct} further divides the links of Table~\ref{tab:deep_coref_head_seen}
based on whether they are correct coreferent links.
The results of Table~\ref{tab:deep_coref_head_seen_correct} show that most of the incorrect links are 
also made between the mentions that are both seen in the training data.

The high ratios indicate that (1) there is a high overlap between the mention pairs of the training and development sets, and
(2) even though that \emph{deep-coref} uses generalized word embeddings instead of exact surface forms, it is strongly biased towards the seen mentions.

We analyze the links that are created by Stanford's rule-based system and
compute the ratio of the links that exist in the training set. 
All corresponding ratios are lower than those of \emph{deep-coref} in Table~\ref{tab:deep_coref_head_seen_correct}.
However, the ratios are surprisingly high for a system that does not use the training data.
This analysis emphasizes the overlap in the CoNLL datasets.
Because of this high overlap, 
it is not easy to assess the generalizability of a coreference resolver to unseen mentions on the CoNLL dataset given its official split.

We also compute the ratios of Table~\ref{tab:deep_coref_head_seen_correct} for the missing links that are associated with the recall errors of \emph{deep-coref}.
We compute the recall errors by cort error analysis tool \cite{martschat14}.  
Table~\ref{tab:deep_coref_recall} shows the corresponding ratios for recall errors.
The lower ratios of Table~\ref{tab:deep_coref_recall} in comparison to those of Table~\ref{tab:deep_coref_head_seen} emphasize the bias of \emph{deep-coref} towards the seen mentions.

For example, 
the \emph{deep-coref} links include 31 cases in which both mentions are either proper names 
or common nouns and the head of one of the mentions is ``country''.
For all these links, ``country'' is linked to a mention that is seen in the training data.  
Therefore, this raises the question how the classifier would perform on a text 
about countries not mentioned in the training data.

Memorizing the pairs in which one of them is a common noun
could help the classifier to capture world knowledge to some extent.
From the seen pairs like (Haiti, his country), and (Guangzhou, the city) 
the classifier could learn that ``Haiti'' is a country and ``Guangzhou'' is a city.
However, it is questionable how useful word knowledge is if it is mainly based on the training data. 
 
%
The coreference relation of two nominal noun phrases with no head match 
can be very hard to resolve.
The resolution of such pairs has been referred to as capturing semantic similarity \cite{clarkkevin16a}.
\emph{deep-coref} links 49 such pairs on the development set.
Among all these links, only 5 pairs are unseen on the training set and all of them are incorrect links.

The effect of lexical features is also analyzed by \newcite{levy15} for tasks like hypernymy and entailment.
They show that state-of-the-art classifiers memorize words from the training data.
The classifiers benefit from this lexical memorization
when there are common words between the training and test sets. 
\section{Discussion}
\label{sect:disc}
We show the extensive use of lexical features biases coreference resolvers 
towards seen mentions.
This bias holds us back from developing more robust and generalizable coreference resolvers.
After all, while coreference resolution is an important step for text understanding, 
it is not an end-task.
Coreference resolvers are going to be used in tasks and domains for which coreference annotated corpora may not be available.
Therefore, generalizability should be brought into attention in developing coreference resolvers. 

Moreover, we show that there is a significant overlap between the training and validation sets in the CoNLL dataset.
The \emph{LEA} metric \cite{moosavi16b} is introduced as an attempt to make coreference evaluations more reliable.
However, in order to ensure valid developments on coreference resolution, 
it is not enough to have reliable evaluation metrics.
The validation set on which the evaluations are performed also needs to be reliable.
A dataset is reliable for evaluations if a considerable improvement on this dataset indicates 
a better solution for the coreference problem instead of a better exploitation of the dataset itself. 

This paper is not intended to argue against the use of lexical features.
Especially, when word embeddings are used as lexical features.
The incorporation of word embeddings is an efficient way for capturing semantic relatedness.
Maybe we should use them more for describing the context and less for describing the mentions themselves. 
Pruning rare lexical features plus incorporating more generalizable features could also 
help to prevent overfitting.

To ensure more meaningful improvements, 
we ask to incorporate out-of-domain evaluations in the current coreference evaluation scheme.
Out-of-domain evaluations could be performed by using either the existing genres of the CoNLL dataset or by using other existing coreference annotated datasets like WikiCoref, MUC or ACE.   

\section*{Acknowledgments} The authors would like to thank Kevin Clark for answering all of our questions regarding deep-coref.
We would also like to thank the three anonymous reviewers for their thoughtful comments. 
This work has been funded by the Klaus Tschira Foundation, Heidelberg,
Germany. The first author has been supported by a Heidelberg Institute for Theoretical Studies
PhD.\ scholarship.

\bibliographystyle{acl_natbib}
\bibliography{../../../bib/lit,mybib}

\end{document}